\documentclass[conference]{IEEEtran}

\usepackage{float}
\usepackage{bm}
\usepackage{tabularx}
\usepackage{longtable}
\usepackage[table,xcdraw]{xcolor}
\usepackage{amsmath,amssymb,amsfonts}
\usepackage{algorithmic}
\usepackage{graphicx}
\usepackage{svg}
\usepackage{textcomp}
\usepackage{booktabs,subcaption,amsfonts,dcolumn}
\usepackage{enumitem}
\usepackage{array}
\usepackage{tcolorbox}
\usepackage{placeins}
\usepackage[]{footmisc}
\usepackage{hyperref}
\usepackage{cleveref}
\hypersetup{
    colorlinks = true,
    linkcolor = black,
    urlcolor  = black,
    citecolor = black,
    anchorcolor = black
}

\newcommand{\changeurlcolor}[1]{\hypersetup{urlcolor=#1}}   

\definecolor{BackgroundBlue}{HTML}{edf2f9}

\definecolor{BorderBlue}{RGB}{15,104,157}

\definecolor{BackgroundGreen}{HTML}{ebf1f1}

\definecolor{BorderGreen}{HTML}{079855}

\newtcolorbox{titleEnvGreen}{
sharp corners,
center,
top=0pt,
width=1.00\linewidth,
halign=center,
valign=center,
colframe=BorderGreen,
colback=BackgroundGreen, 
boxrule=0pt,
rightrule=0pt,
bottomrule=0pt,
leftrule=4pt
}

\newtcolorbox{titleEnvBlue}{
sharp corners,
top=4pt,
left=0pt,
right=0pt,
bottom=6pt,
width=1.00\linewidth,
halign=center,
valign=center,
colframe=BorderBlue,
colback=BackgroundBlue, 
boxrule=0pt,
rightrule=0pt,
bottomrule=0pt,
leftrule=4pt
}

\begin{document}

\title{A Roadside Unit for Infrastructure Assisted Intersection Control of Autonomous Vehicles}

\author{Michael Evans$^{1,*}$, Marcial Machado$^{2}$, Rickey Johnson$^{3}$, Anna Vadella$^{4}$, Luis Escamilla$^{5}$, \\ Beñat Froemming-Aldanondo$^{6}$, Tatiana Rastoskueva$^{7}$, Milan Jostes$^{8}$, Devson Butani$^{8}$,\\ Ryan Kaddis$^{8}$, Chan-Jin Chung$^{8}$, and Joshua Siegel$^{9}$%

\\\\

\small{$^{1}$Old Dominion University,} \small{$^{2}$The Ohio State University,} \small{$^{3}$North Carolina A\&T State University,}\\ \small{$^{4}$Butler University,} \small{$^{5}$New Mexico State University,} \small{$^{6}$University of Minnesota,}\\ \small{$^{7}$The University of Arizona,} \small{$^{8}$Lawrence Technological University,} \small{$^{9}$Michigan State University}\\ \small{$^{*}$Corresponding Author: mevan028@odu.edu}}

\maketitle

\begin{abstract}

Recent advances in autonomous vehicle technologies and cellular network speeds motivate developments in vehicle-to-everything (V2X) communications. Enhanced road safety features and improved fuel efficiency are some of the motivations behind V2X for future transportation systems. Adaptive intersection control systems have considerable potential to achieve these goals by minimizing idle times and predicting short-term future traffic conditions. Integrating V2X into traffic management systems introduces the infrastructure necessary to make roads safer for all users and initiates the shift towards more intelligent and connected cities. To demonstrate our control algorithm, we implement both a simulated and real-world representation of a 4-way intersection and crosswalk scenario with 2 self-driving electric vehicles, a roadside unit (RSU), and a traffic light. Our architecture reduces acceleration and braking through intersections by up to 75.35\%, which has been shown to minimize fuel consumption in gas vehicles. We propose a cost-effective solution to intelligent and connected intersection control to serve as a proof-of-concept model suitable as the basis for continued research and development. Code for this project is available at \changeurlcolor{magenta}{\url{https://github.com/MMachado05/REU-2024}}.

\end{abstract}


\section{Introduction}
Crosswalks oriented along street intersections are among the most dangerous for pedestrians, accounting for up to 60\% of injuries caused by vehicles in cities such as Montreal \cite{brosseau2013impact}. This vulnerability demands a safer approach to managing traffic; one method of achieving safer crosswalks is to deploy V2X-enabled roadside units for adaptive intersection control. \par

Vehicle-to-vehicle (V2V) \cite{siegel2017survey}, vehicle-to-cyclist~\cite{9305274}, and vehicle-to-infrastructure (V2I) communications \cite{santa2008architecture} are incorporated into modern vehicles to optimize for functions such as avoiding delays or minimizing stop counts \cite{placzek2012self,siegel2018algorithms}. V2I optimization methods are categorized as NP-hard problems \cite{giridhar2006scheduling}, limiting the scope of usability to centralized data due to the algorithmic complexity of the computations \cite{placzek2012self}. Similar V2X approaches to scheduling optimization \cite{li2017recasting} use real time traffic information instead of centralized data and can enable drivers to take early action \cite{huang2020recent}. Dedicated Short-Range Communication (DSRC) radios \cite{kenney2011dedicated} have long been used for V2V communication \cite{zhang2018virtual}, and are utilized in research implementations of safety features such as crash avoidance. By aggregating existing technologies from DSRC and cellular networks, V2X offers a software framework for exchanging information between vehicles and components of the intelligent transportation system (ITS) \cite{abboud2016interworking}. To enhance the accessibility of our V2X learning tool, we utilize Wi-Fi as an alternative to conventional cmWave technologies, such as DSRC or ITS-G5. \par

We propose a low-cost wireless intersection control architecture with an RSU as a teaching tool to demonstrate potential improvements to the metrics of energy consumption, safety, and driving behaviors enabled by connected vehicles. Our intersection and crosswalk scenario model is implemented using two Polaris Gem e2 electric vehicles (EVs) known as Autonomous Campus Transport (ACTor) vehicles. Each ACTor is equipped with the hardware necessary for self-driving, including a Dataspeed Drive-by-Wire kit, HDR Camera, 3D LiDAR sensors, two Swift Piksi GPS modules, and a computer for programming the Drive-by-Wire system with ROS \cite{quigley2009ros}. \par

\begin{figure}[h!]
\centering
\includegraphics[width=0.849\linewidth]{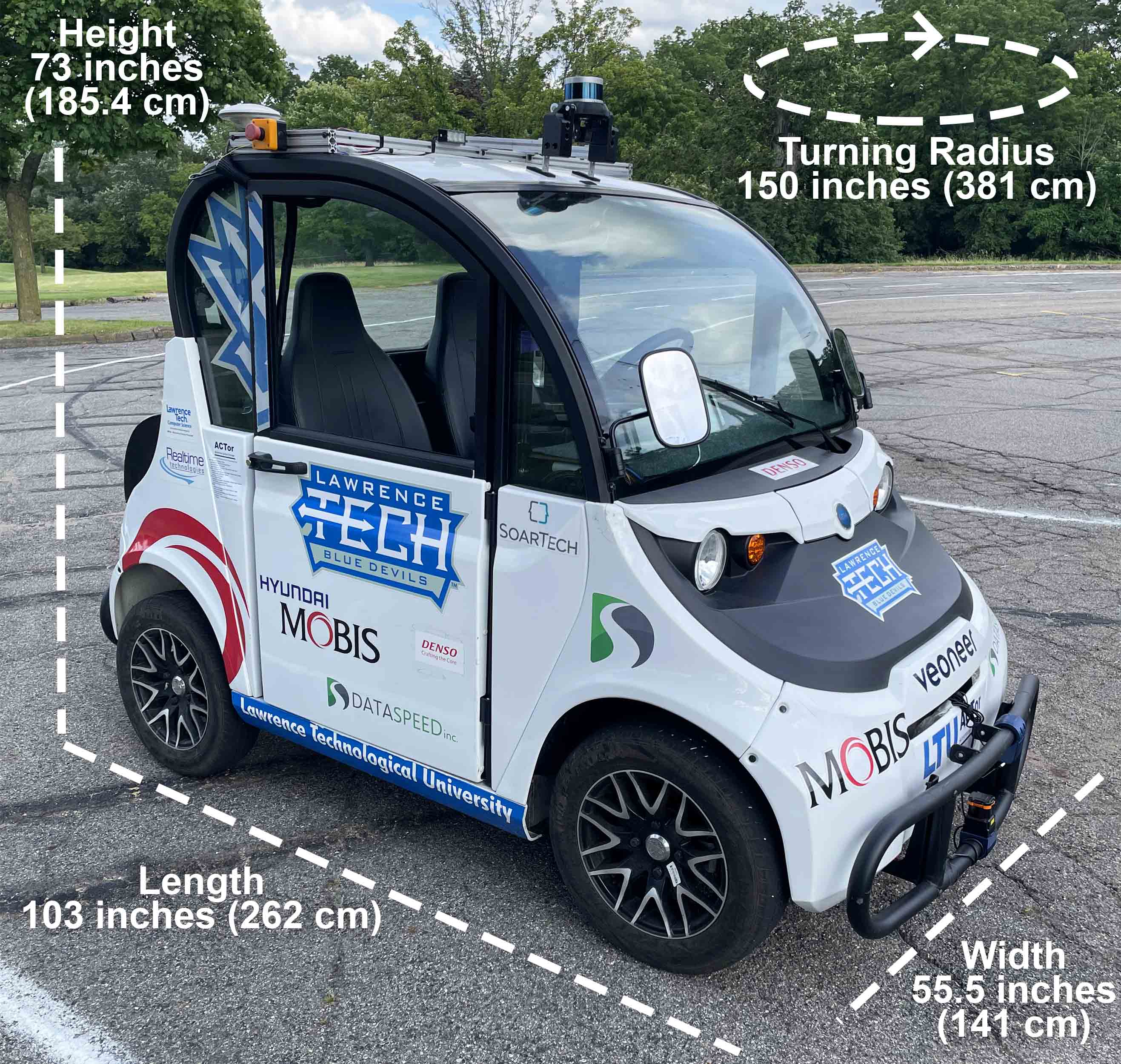}
\caption{Image of the Polaris GEM e2 ``ACTor 1'' vehicle.}
\label{fig:ACTor1}
\end{figure}

To make the progression towards adaptive intersections for autonomous vehicles, we introduce a V2X proof-of-concept system consisting of an RSU, 2 self-driving electric vehicles, and an Arduino-powered traffic light for state visualization. The significant contributions of our research, outlined in the remainder of the paper, are as follows:
\begin{enumerate}
  \item Simulate our intersection control system first in a virtual environment of our Lot H test course using GazelleSim.
  \item Establish a V2I connection between the ACTors and RSU to enable vehicle speed control within the intersection.
  \item Develop and evaluate a real-world representation of our V2X teaching tool operating in both an emulated 4-way intersection and cross-walk scenario.
\end{enumerate}

\section{Review of Literature}
\label{sec:lit-review}
Most existing research efforts on infrastructure-assisted intersection control are only tested in simulation \cite{soto2022survey, choi2020qos} due to the costs and logistics of real-world models for such implementations. State-of-the-art projects combine network simulators such as ns-3 \cite{campanile2020computer} with traffic simulators like SUMO \cite{kusari2022enhancing} for applications on vehicle platooning, sensor sharing, and communication protocol conformance testing \cite{s19020334}. While these projects aim to provide realistic V2X scenarios by running robust simulation, their pursuits are still not perfect representations of the physical environments they aim to model. Communication simulation delay is introduced in virtual environments, which, if high, can not reflect reality \cite{s19020334}. Obstacles such as adverse weather conditions, variances in the kinematics between vehicle types, uneven pavement conditions, driver ability \cite{hosseinlou2012study}, and networking limitations can not be understood by simulation alone. The speed of these simulations is also a concern, as performance degrades exponentially when modeling high traffic densities \cite{ORMANDI2024103003}. \par

Gaps in the research also exist for work that includes field testing. Lu, Jung, and Kim \cite{lu2021optimization} propose a solution called \textit{Vehicle-to-Intersection}, which assumes all vehicles are capable of autonomous collision avoidance and are controlled by an Intersection Control Agent, however, they do not consider deployment costs. An approach for displaying occupancy grids of vehicles in close proximity with an RSU has been covered in \cite{khalfin2023vehicle}, but does not provide the autonomous intersection control that our research emphasizes. Specific use cases for intersection control are proposed in \cite{le2009v2x} for crossing-path collision avoidance by broadcasting sensor information about other road users at set frequencies. Efforts to lower the cost of deployment exist by reducing the number of units required through optimizing coverage ratios \cite{ben2021roadside}, or supplementing parked cars as RSUs \cite{reis2017parked}, but are bound by deployment specifications which our teaching model does not consider.

\par

Other works considered in this section examine specific and important components of connected traffic management with V2X. The existing gaps in research are due to using fully simulated environments, not considering deployment costs, or requiring human input in response to an in-car display. Our research contributes to this area by incorporating an adaptive-speed algorithm into a connected intersection enabled by communication with a cost-efficient roadside unit. 

\begin{figure}[h]
\centering
\includegraphics[width=0.87\linewidth]{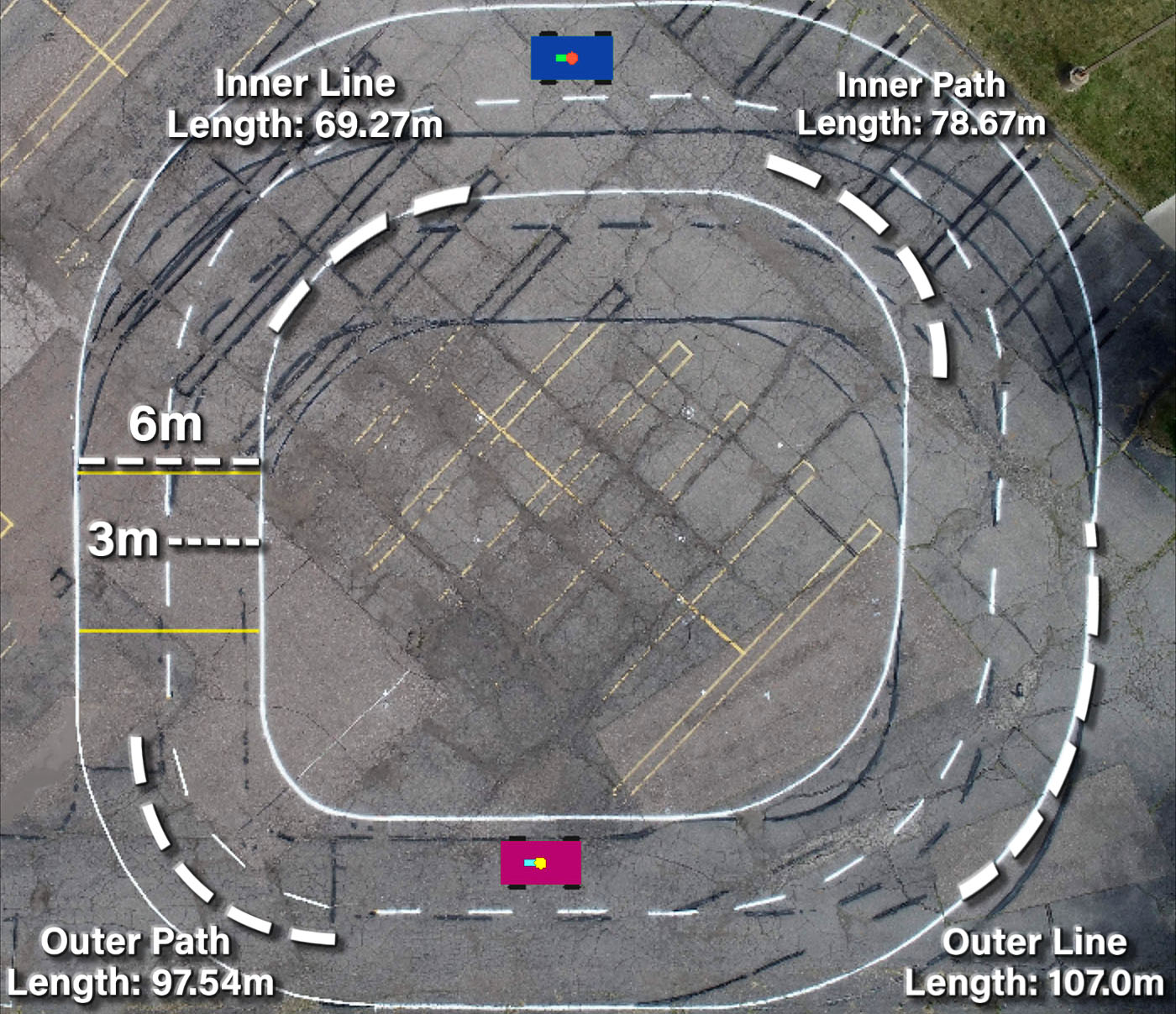}
\caption{Map of the Lot H test course in GazelleSim.}
\label{fig:Gazelle}
\end{figure}

\section{System Design and Methods}
To evaluate the performance of our infrastructure-assisted control algorithm, we first develop a virtual simulation using GazelleSim \cite{pappas2021gamified}, and an aerial-view mapping shown in Figure \ref{fig:Gazelle} of the Lawrence Technological University (LTU) Parking Lot H located in Southfield, Michigan. This simulator is used for its ability to simulate multiple agents simultaneously, and lightweight power consumption, enabling simulation testing on field laptops. Next, we define the implementations of the real-time system including the RSU, ACTor vehicles, and traffic light for human visualization. An overview of the ROS software architecture components and detailed summary of each respective development follows. 

\subsection{Gazelle Simulation}
As suggested in Section \ref{sec:lit-review}, it is not always practical or possible to acquire the hardware needed for real-world testing of a V2X system. In such situations, or for initial testing of new software features, algorithms are first implemented in simulation. This method of software development allows for the debugging of new features in a controlled, safe, and weather-independent environment. We use a lightweight simulator developed at LTU: GazelleSim, two Ackermann steering robots with turning capabilities identical to the ACTor vehicles, and an aerial photo of the Lot H course as our simulation environment. The simulator uses a \textit{meters per pixel} parameter to accurately display the position of both vehicles on the map as they would appear in a real-world test at that GPS location. By combining the meters per pixel parameter with a starting point GPS coordinate plotted on the map, our simulated environment accurately represents the real-world location both visually and geographically. \par

Each vehicle follows a software architecture similar to Section \ref{sec:software-architecture}, using software-in-the-loop (SIL) testing for the Drive-by-Wire system and GPS which are simulated and handled by GazelleSim. In this virtual environment, the vehicle's controller publishes twist messages in the form of linear and angular x, y, and z values instead of native Drive-by-Wire commands. All simulated agents have access to their x and y coordinates for continuous tracking position without requiring latitude and longitude from a GPS node. Another deviation from real-world testing is that the RSU is implemented as a ROS node: an efficient virtualization of the physical Ubuntu server and network. RSUs can be exorbitant to implement \cite{al2018towards}, often necessitating the use of virtual testing before implementing the hardware and networking required of a physical system. The remainder of the simulation aligns with the real-world implementation as discussed in Section \ref{sec:software-architecture}.


\subsection{Software Architecture}
\label{sec:software-architecture}
We develop a high-level software architecture and a corresponding ROS workspace directory structure to increase modularity and ease of iteration. Our codebase is organized into three primary categories: the hardware interface, drive logic, and traffic light model. This modular structure allows for the efficient substitution of discrete code sectors, such as the lane detection node, for making exploratory system performance comparisons with minimal overhead. \par

Our hardware-interfacing logic handles three primary functions: a package for reading image information from the ACTor's HDR camera, one that receives GPS information from the Piksi GPS nodes, and an API for communicating movement directives to the ACTor vehicles. These packages function as a mediator between physical objects and the data they capture. The next section will outline the responsibility of each software node in the system architecture. \par

We use an image preprocessor node to receive raw images from the camera-compatible package and prepare them for lane detection through processing algorithms such as median blur, canny edge detection, and white filtering. This node provides the functionality of altering the raw image separately from the lane following logic, as some algorithms perform better with differing parameters. Without this, various lane-following algorithms will need to process the image individually, causing large repetitions in the code. Next, the processed image is input into the lane detection node which calculates the locations of lane borders to derive a desired turn angle. \par

With lane following handled by the previous steps, our adaptive speed algorithm logic will now be explained. The vehicle's current location and internally saved course waypoints are used to calculate the distance to the intersection from either lane. Further information regarding this calculation are explained in Section \ref{sec:adaptive-speed-control}. The code implemented on the RSU, which is activated via an SSH connection into the Raspberry Pi at runtime, works alongside the distance-to-intersection node to manage the intersection. We recognize the security risks associated with SSH and emphasize that our model is intended as a teaching tool, demonstrating V2X concepts rather than serving as a deployable solution to intersection control. On the RSU is the traffic light algorithm, which keeps track of an abstraction of its current light color, and the duration of all states. Each traffic light state consists of a configurable duration and a binary representation of either a red or green light. This node publishes information on state changes of the traffic light, the time left in the current state, and the duration of the next state in three unique ROS topics. Each topic is subscribed to by the vehicle controller node, which, together with distance calculation, allows for our adaptive speed algorithm to function. The RSU's primary function is to publish information about the traffic light to provide intersection awareness to the vehicles, which individually alter their speeds to avoid red lights in response, if required. \par

Finally, the vehicle controller node takes information from the distance-to-intersection node and the lane detection node to make a concluding decision on vehicle movement. All other nodes serve in some capacity as input parameters to the final output command from the vehicle controller.

\subsection{RSU Hardware and Traffic Light Visualization}
To generate a V2I connection between the vehicles and RSU, we use a 12V battery, 12V-5V converter, a Raspberry Pi 4 B single-board computer running the Ubuntu Server 20.04 operating system, and an external Netgear router with range extenders and a wireless access point. Our traffic light is composed of 4 individual LED lights, relays, a controller for the relays, a power supply, a logic radio module/wifi, and is controlled through an Arduino Wemos D1 board. On state change, the RSU sends a Rosserial message with the state of the traffic lights to the board. This message in turn sets the pins of the traffic light to light up the specific light configuration to show the respective state. As connected vehicles have access to the duration of the light, yellow lights are unnecessary and act as a fail-operational identifier for a loss of connection. 

\subsection{Adaptive Speed Control}
\label{sec:adaptive-speed-control}
To reduce vehicle emissions and noise pollution attributed to idling at red lights \cite{mahler2012reducing, tiwari2013fuel}, the RSU sends data to the vehicle controller which adjusts the vehicle's speeds. In effect, this synchronizes each agent's approach to the intersection with the green light timings. Variations in vehicle speeds at traffic intersections lead to an increase in fuel usage and a decrease in air quality for the immediate and surrounding area \cite{li2018separation}. To solve this, we construct an adaptive speed algorithm using GPS waypoints and a kinematics equation to calculate the target average velocity that each vehicle should drive to arrive at an intersection as the traffic state switches from red to green. \par

Our process for capturing the waypoints involves pre-recording GPS coordinates of the test course and the intersection in lieu of commercially available HD automotive maps data. During this process, one student pilots the ACTor while another indicates where to record each waypoint for a consistent distance of approximately 3 meters between measurements. The latitude and longitude values of each intersection and standard waypoint are saved to 2 yaml files. \par

At each light state change, the vehicle controller determines if each vehicle can cross the intersection driving at its current velocity before the traffic state changes again. If so, or if the vehicle loses connection to the RSU, the vehicle controller assigns no change in target velocity. If it can not make the intersection in time, the following process is used to calculate an average target velocity to maintain to prevent stopping. \par

\begin{titleEnvBlue}
\noindent
\textcolor{BorderBlue}{\textbf{Definitions:}}
\begin{enumerate}[label={},ref=\Alph*,leftmargin=*, itemsep=1pt]
  \item Let \( v = (\phi_v, \lambda_v) \) be the coordinates of vehicle \( v \).
  \item Let \( w_i = (\phi_i, \lambda_i) \) be the coordinates of waypoint \( w_i \).
  \item Let \( W = \{w_1, w_2, \ldots, w_n\} \) be the set of all waypoints.
  \item Let \( I \subset W \) be the set of all intersections.
\end{enumerate}
\end{titleEnvBlue}
\noindent
  The distance between 2 points on a sphere can be calculated using the Cosine-Haversine formula \cite{robusto1957cosine} \( h(\phi_1, \lambda_1, \phi_2, \lambda_2)= \)

  \begin{equation}
  {2r \cdot \arcsin \sqrt{\sin^2\left(\frac{\Delta\phi}{2}\right) + \cos(\phi_1) \cos(\phi_2) \sin^2\left(\frac{\Delta\lambda}{2}\right)}}
  \end{equation}

\noindent
\text{Find the waypoint \(w_i\) with the shortest distance to \( v \):}

  \begin{equation}
  \text{Let } p = \min_{x \in W} h(x, v)
  \end{equation}

\noindent
\text{Sum \( h(w_i, w_{i+1}) \) from \( p \) to the next intersection:}

  \begin{equation}
  \Delta x_{\text{total}} = \sum_{i=p}^{k-1} h(w_i, w_{i+1}) \quad : w_k \in I \text{ and } k > p
  \end{equation}

\noindent
\text{Find the target average velocity the vehicle should drive:}

\begin{equation}
v_{\text{avg}} = \frac{\Delta x}{\Delta t} \quad : \Delta x = \Delta x_{\text{total}} \text{ and } \Delta t = \text{light duration}
\end{equation}

\begin{figure}
\centering
\begin{subfigure}{0.50\textwidth}
   \includegraphics[width=1\linewidth]{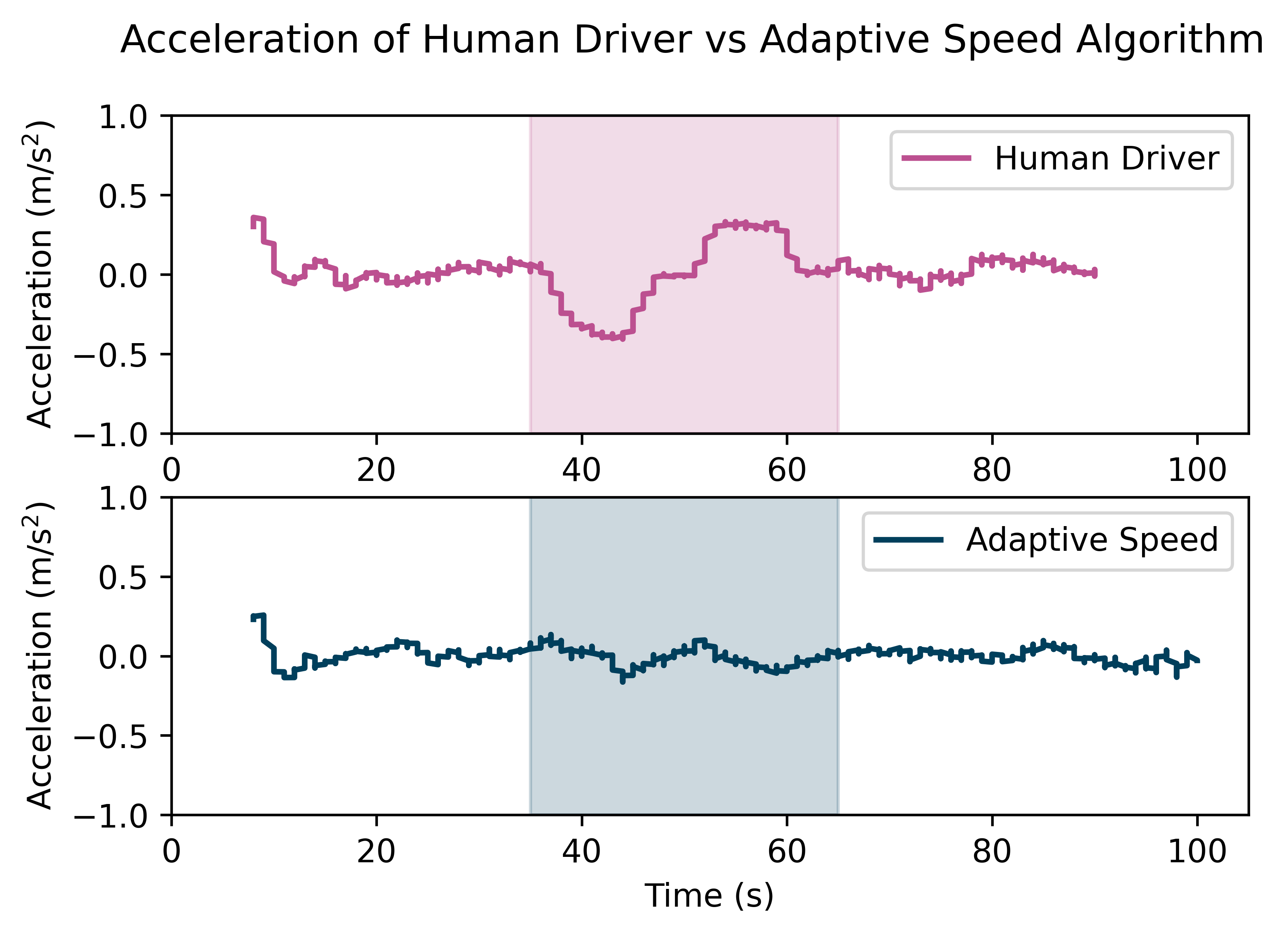}
   \caption{40s green / 10s red light state}
   \label{fig:Ng1} 
\end{subfigure}

\begin{subfigure}{0.50\textwidth}
   \includegraphics[width=1\linewidth]{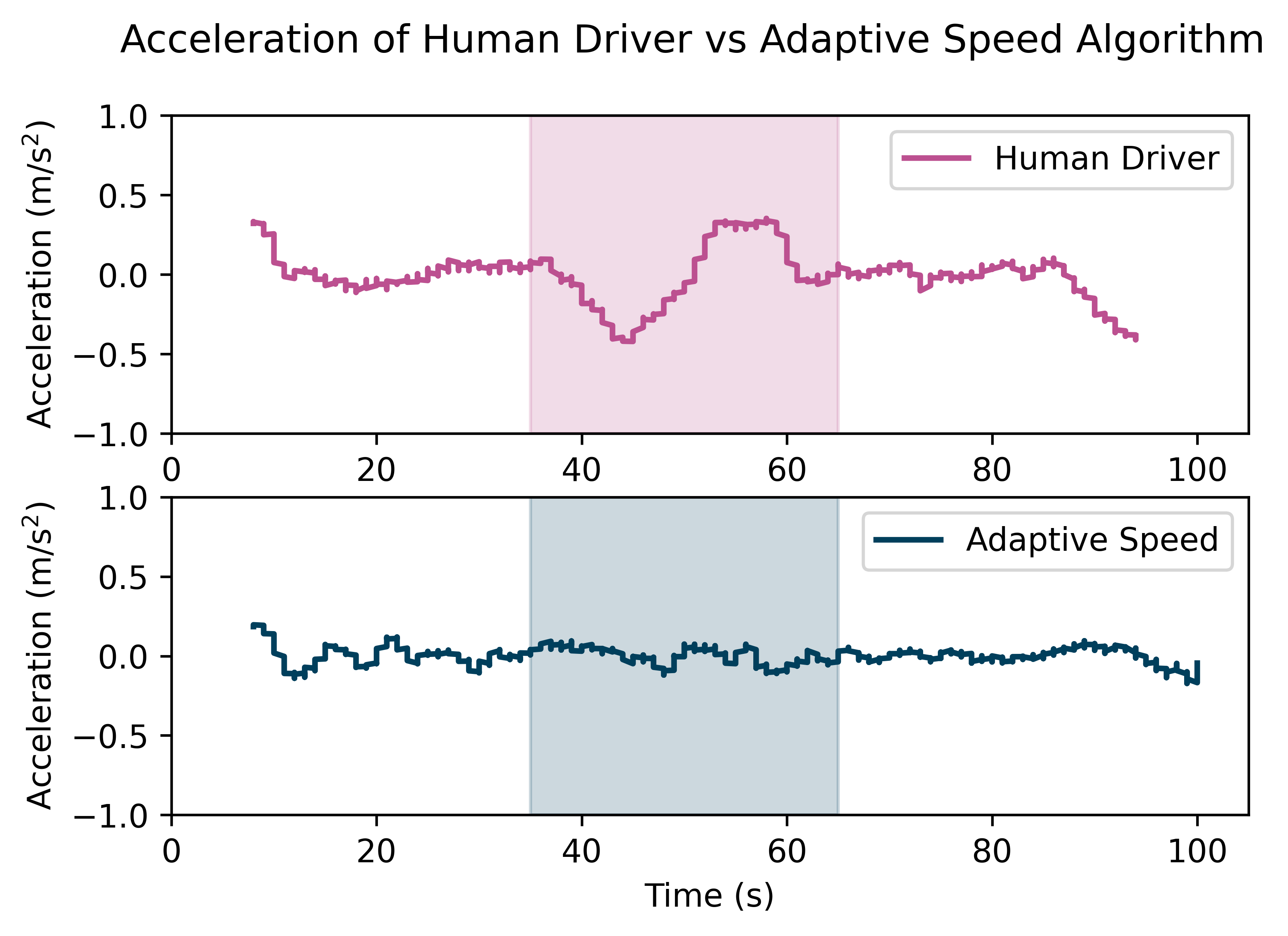}
   \caption{25s green / 25s red light state}
   \label{fig:Ng2}
\end{subfigure}

\begin{subfigure}{0.50\textwidth}
   \includegraphics[width=1\linewidth]{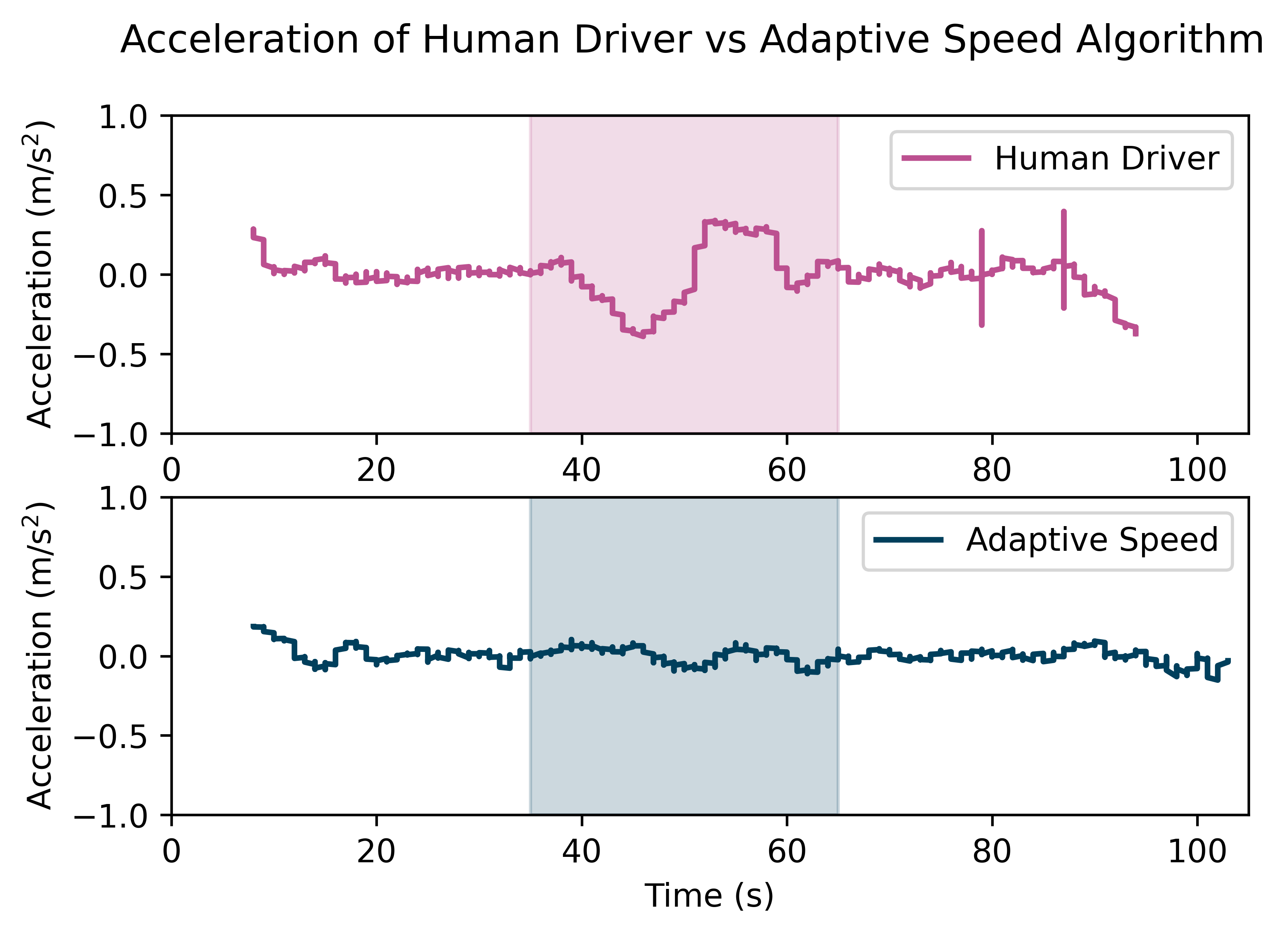}
   \caption{10s green / 40s red light state}
   \label{fig:10G40RAcc} 
\end{subfigure}

\caption{Acceleration vs time comparison of a human driver against our adaptive speed algorithms during 3 different light state configurations. The highlighted region is from 35s to 65s and captures the vehicle crossing the intersection.}
\label{fig:acceleration-vs-time}
\end{figure}

\section{Cost-Effectiveness and Scalability}
As the required RSU density for connected vehicles scales with traffic density \cite{degrande2021c}, analyzing the cost efficacy of each unit has become a valuable benchmark in this field. Cost of ownership must consider factors such as coverage, location-dependent installation costs, energy consumption, and hardware expenses \cite{degrande2021c}. However, our focus is on developing a V2X teaching tool within a constrained space, so we limit our cost analysis to hardware expenditures. The real estimated deployment costs for an RSU can range from \$7,000 to \$15,000 \cite{ITS_America_V2X_Deployment_Plan_2023}, with a more comparable cost exclusive to hardware being €6000 ($\sim$\$6300) \cite{degrande2021c}. By utilizing simplified communication protocols and limiting coverage requirements, our proposed RSU can be deployed in a teaching environment for \$200.  \par

Through deliberately abstracting away from many of the intricacies of full-scale deployment, our platform enables learners to focus on the fundamental aspects of algorithm design and performance analysis. This simplified environment creates opportunities for rapid prototyping and testing, encouraging active exploration of the design space and facilitating the discovery of novel and effective solutions, unencumbered by extraneous operational complexities. As a result, we anticipate compute limitations to become a concern as the number of vehicles managed by the same RSU surpasses 10. Furthermore, the compute required to adjust for significant differences in stopping distances or vehicle fuel types may introduce additional concerns for scalability with our current hardware.

\begin{figure*}[h]
    \centering
    \subfloat[\centering 40s green / 10s red light state]{{\includegraphics[width=0.2731\textwidth]{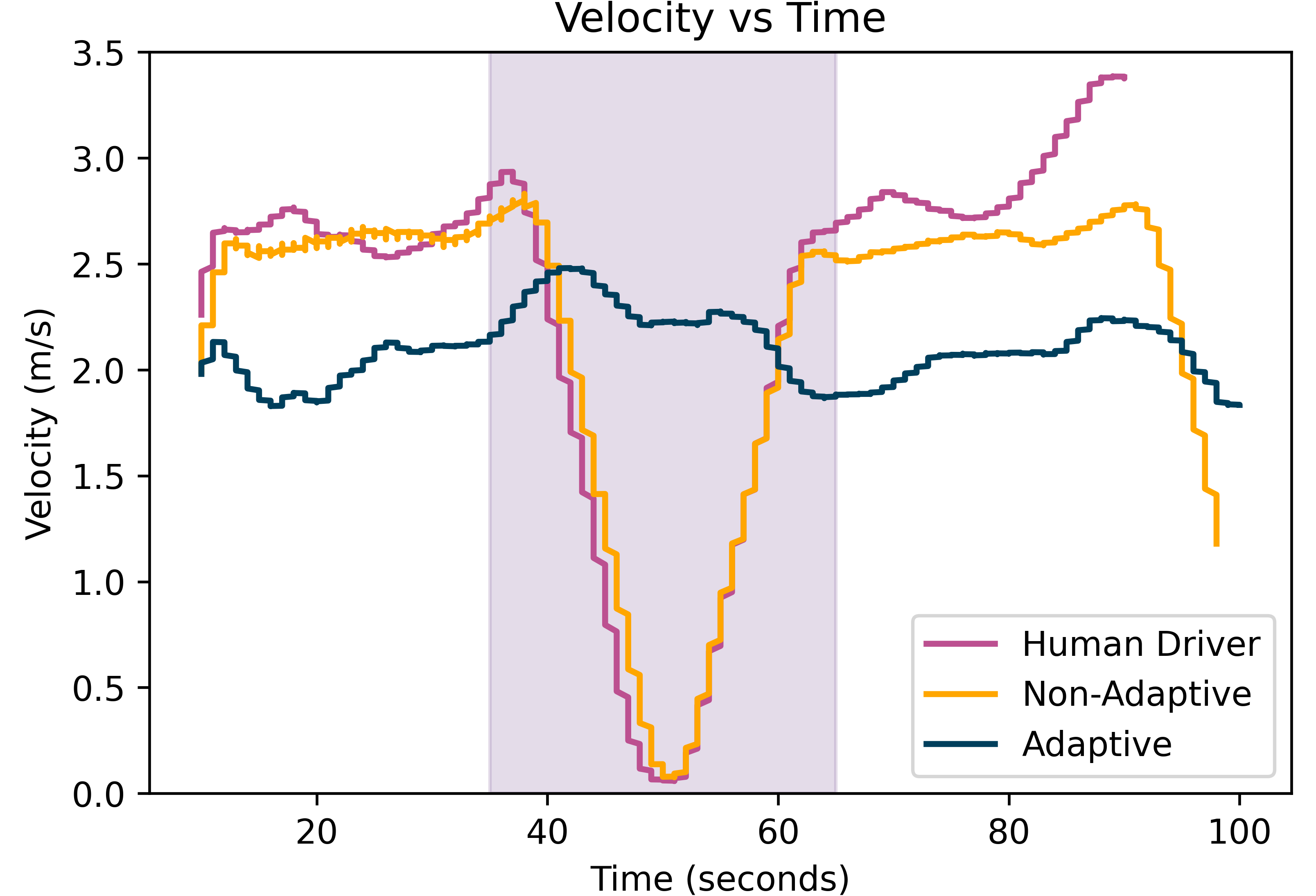} }}%
    \qquad
    \subfloat[\centering 25s green / 25s red light state]{{\includegraphics[width=0.2731\textwidth]{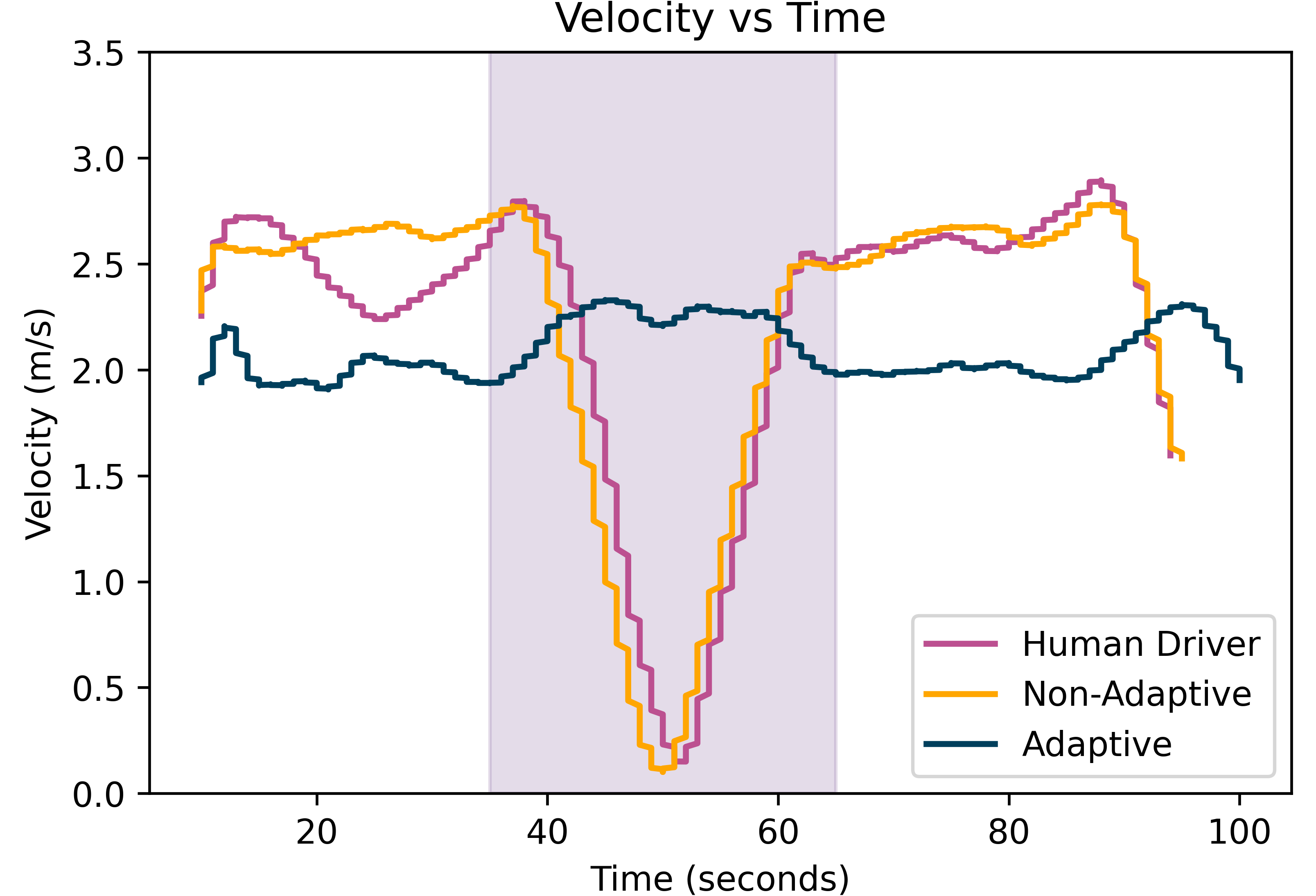} }}%
    \qquad
    \subfloat[\centering 10s green / 40s red light state]{{\includegraphics[width=0.2731\textwidth]{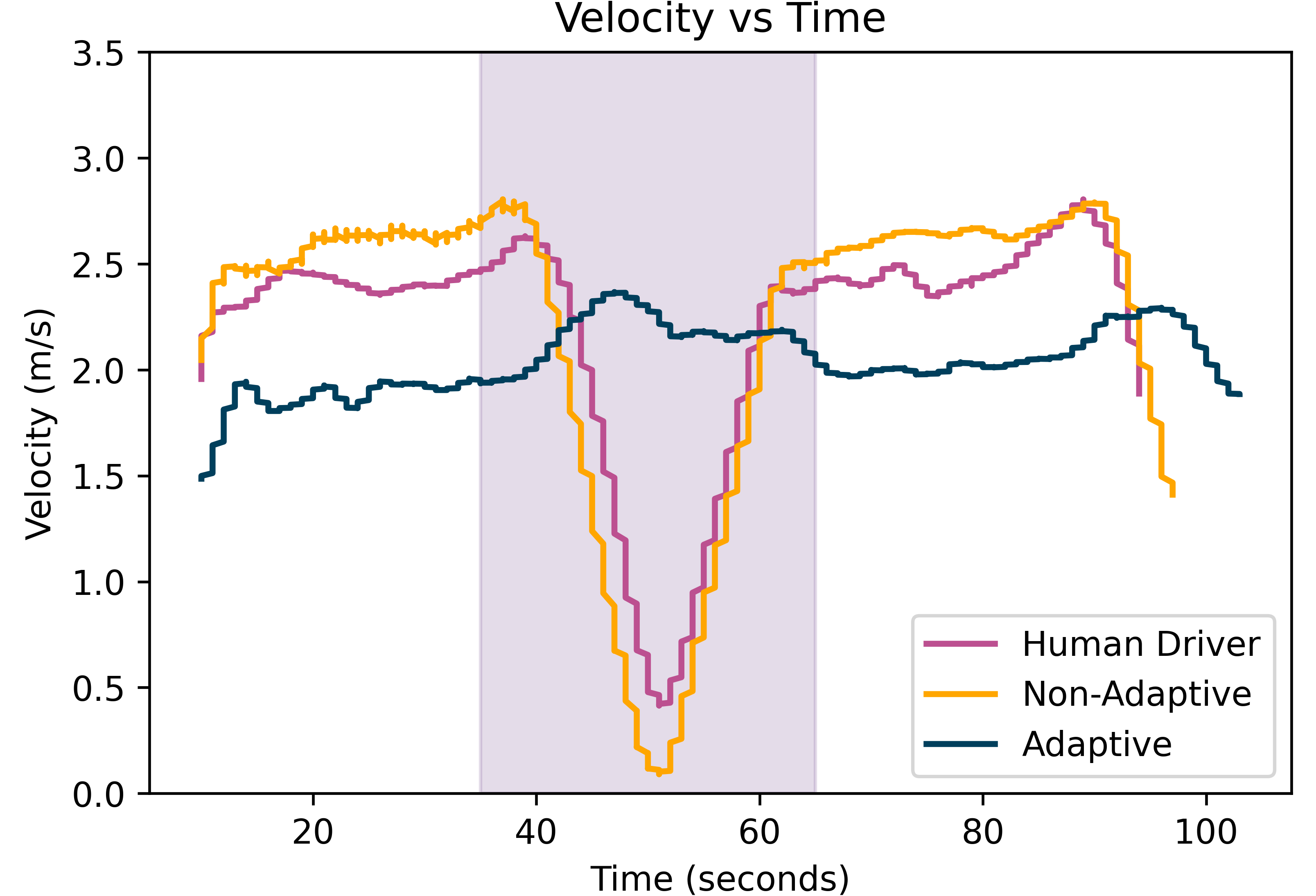} }}%
    \caption{Velocity vs time comparison of all driving methods against 3 light state configurations.}%
    \label{fig:velocity-vs-time}
\end{figure*}

\section{Experiment and Results}
In this section, we introduce the real-world environment for testing our adaptive speed algorithm on two electric vehicles. Then, we compare the average velocity and acceleration of the vehicles around and through the intersection for each task. Two scenarios are designed for this project: a crosswalk with synchronized light states allowing pedestrians to pass a critical zone, and a simplified 4-way intersection with no left turns, and independent light states. Finally, we analyze our results to show that our adaptive speed algorithm reduces the total change in speed through the intersection, and prevents idling. \par

\subsection{Experimental Setup}
Data collection for each evaluation begins at the initial green light encountered by the ACTor, which is enabled to drive the first frame that a green light is registered. For each experiment, we also include a human driver to serve as a control. The driver is instructed to maintain a speed of 5 mph (2.24 m/s) and receives a verbal 5-second warning of state changes to emulate real-world yellow lights. Additionally, each experiment is finalized as the ACTor encounters the intersection in the outer lane for a second time, noted by the top yellow line in Figure \ref{fig:Gazelle}. We also capture z axis angular velocity with the vehicle's IMU for an evaluation on passenger comfort in self-driving vehicles, detailed in Figure 8 of \cite{evans2024vehicle}.


\subsection{Performance Analysis}
We evaluate this V2X architecture both quantitatively and qualitatively in the form of velocity, cost to deploy in a teaching environment, and reductions made to acceleration across the intersection. The isolated range of values in the highlighted region of Figure \ref{fig:acceleration-vs-time} is used to analyze the percent decrease in vehicle speed-ups and braking through an intersection.

\begin{equation}
\left( \frac{\int_{t=35s}^{t=65s} |a(human)|dt - \int_{t=35s}^{t=65s} |a(adaptive)|dt}{\int_{t=35s}^{t=65s} |a(human)|dt} \right) \cdot 100
\end{equation}

Our proposed algorithm reduces the total change in velocity through the highlighted region of 73.15\%, 75.35\%, and 73.79\% respectively across the three trials when compared to a human driver. By minimizing changes in velocity at intersections, we expect a reduction in fuel consumption in gas vehicles \cite{almannaa2017reducing, stevanovic2021fuel}. Moreover, this non-deployable architecture significantly reduces costs, with each unit requiring \textless5\% of the anticipated hardware expenses for deployable units. Minimal overhead may increase V2X exposure and lead to the widespread use of pedestrian detection systems, providing an earlier warning than conventional mechanisms \cite{arena2019v2x, larue2020pedestrians}. \par

\begin{table}[h!]
    \centering\small
   \caption{\small Comparison of the average velocity of the V2X algorithms against a human driver during 2 laps around our test course.
 \label{tab:consistency}}
    \begin{tabular}{c|c|c|c}
    \toprule
    {\bf Light States} & {\bf Human} &  {\bf Non-Adaptive} & {\bf Adaptive}\\
    \midrule
    40s green / 10s red & 2.34 m/s & 2.10 m/s & 2.07 m/s \\
    25s green / 25s red & 2.14 m/s & 2.16 m/s & 2.06 m/s \\
    10s green / 40s red & 2.14 m/s & 2.18 m/s & 1.95 m/s \\
    \bottomrule
    \end{tabular}
    \label{tab:v2x-metrics}
\end{table}

Table \ref{tab:v2x-metrics} also shows a comparison of average velocity for a more quantitative evaluation of Figure \ref{fig:velocity-vs-time}. One possible bias that we acknowledge is the duration of the light states. Configurations that do not force a red light encounter on intersection approach could generate less conclusive results. Another potential bias is vehicle speeds and traffic density. Our system is only tested within the capabilities of the lane-following algorithms, unfit for high speeds or dense traffic. These are two known network complications present in cities. \par


\section{Conclusion and Future Work}
In this paper, we propose a cost-effective V2X teaching model for adaptive intersection control that eliminates red light idling, and reduces the total acceleration and braking through intersections by up to 75.35\%. Our approach utilizes a cost-efficient roadside unit for wireless communication between vehicles and the traffic light state for a more fuel efficient intersection. Another limitation to address as an extension to this research is implementing additional fail states into our system. Constructing a fail-operational method during a loss of connection not reliant on human intervention or full stops would be ideal. Continued functionality during system failure is a requisite safety feature for deployment in uncontrolled city environments, such as intersections and crosswalks.

\section{Acknowledgment}
This project is supported by the National Science Foundation
REU Site Awards \#2150096 and \#2150292. The electric vehicles used for this research are sponsored by MOBIS, US Army GVSC, NDIA, DENSO, SoarTech, Realtime Technologies, Veoneer, Dataspeed, GLS\&T, and LTU.

\bibliographystyle{IEEEtran}
\bibliography{refs}

\end{document}